\documentclass[conference]{IEEEtran}
\IEEEoverridecommandlockouts

\usepackage{cite}
\usepackage{amsmath,amssymb,amsfonts}
\usepackage{algorithmic}
\usepackage{graphicx}
\usepackage{textcomp}
\usepackage{xcolor}

\usepackage{booktabs}
\usepackage{multirow}
\usepackage{xspace}
\newcommand{\method}{PEAR-MIL\xspace}
\def\BibTeX{{\rm B\kern-.05em{\sc i\kern-.025em b}\kern-.08em
    T\kern-.1667em\lower.7ex\hbox{E}\kern-.125emX}}
\begin{document}

\title{PEAR-MIL: Prototype Evidence Aggregation and Retrieval Multi-Instance Learning for Whole Slide Image Classification
\thanks{This work was supported by the National Natural Science Foundation of China under Grant No. 62506025.}
\thanks{\textsuperscript{*}Corresponding author: Qingjie Liu (email: qingjie.liu@buaa.edu.cn).}
}

\author{\IEEEauthorblockN{Shiwei Li\textsuperscript{1}, Lu Wang\textsuperscript{2},
Xinyue Liu\textsuperscript{3}, Huaguan Chen\textsuperscript{3},\\
Xinyi Li\textsuperscript{4}, Qu He\textsuperscript{4}, Qingjie Liu\textsuperscript{3,*}}
\IEEEauthorblockA{\textsuperscript{1}\textit{Hangzhou Innovation Institute, Beihang University, Hangzhou, China}\\
\textsuperscript{2}\textit{Department of Oncology, Tongji Hospital, Tongji Medical College,}\\
\textit{Huazhong University of Science and Technology, Wuhan, China}\\
\textsuperscript{3}\textit{State Key Laboratory of Virtual Reality Technology and Systems,}\\
\textit{Beihang University, Beijing, China}\\
\textsuperscript{4}\textit{School of Computer Science and Engineering, Beihang University, Beijing, China}\\
\{shiweili93, xinyueliu, chg12135, lixinyi2023, 23230608, qingjie.liu\}@buaa.edu.cn\\
wanglu@hust.edu.cn}
}

\maketitle

\begin{abstract}
Weakly supervised whole-slide image (WSI) classification is commonly formulated as multiple instance learning (MIL), where slide-level labels supervise the aggregation of thousands of patch features. Attention pooling and prototype-guided aggregation offer different ways to summarize patch evidence. We investigate whether attention-guided class and background prototypes can retrieve evidence that complements an attention-derived slide representation. To this end, we propose PEAR-MIL, a Prototype Evidence Aggregation and Retrieval Multi-Instance Learning framework. PEAR-MIL jointly learns a compact set of class- and background-associated prototypes via an attention-guided objective, and assigns them a distinct functional role: each prototype serves as a slide-internal evidence query that retrieves a prototype-specific evidence token from the WSI bag through similarity-based matching. An evidence router then aggregates these tokens, and the resulting prototype-indexed evidence is fused with an attention-derived slide feature to provide complementary evidence views for classification. PEAR-MIL operates purely on visual patch features, requires only slide-level labels, and adds minimal parameters beyond a standard attention-based MIL backbone. Experiments on Camelyon16, Camelyon17, and TCGA-NSCLC across multiple feature extractors show that PEAR-MIL achieves the best AUC in several settings and remains competitive in saturated feature regimes.
\end{abstract}


\section{Introduction}
\label{sec:intro}

Whole-slide images (WSIs) capture tissue morphology at gigapixel scale, but detailed region annotations require costly expert review.
Multiple instance learning (MIL) addresses this challenge by treating each slide as a bag of patch features supervised by a slide-level label \cite{campanella2019clinical,lu2021CLAM,ilse2018attention}.
Because diagnostic tissue may occupy only a small fraction of a slide, the aggregator must select and combine evidence from thousands of patches.

Modern WSI pipelines use frozen feature extractors followed by trainable MIL aggregators.
Attention weighting, relation modeling, and prototype-based grouping offer different ways to organize patch evidence \cite{lu2021CLAM,shao2021transmil,zhang2024acmil,tang2023mhim,yang2023tpmil}.
An attention-derived slide feature provides a weighted summary of the bag, while prototype-indexed retrieval can expose additional evidence views associated with recurring feature patterns.
We investigate how these two forms of evidence can be combined for slide classification.

We propose \method, a dual-evidence MIL framework operating on a shared set of projected patch features.
The attention path produces a gated-attention slide representation; the prototype path uses learned class- and background-associated prototypes as conditional evidence queries.
Each query retrieves a prototype-specific evidence token through similarity-based matching, and an evidence router aggregates the tokens.
The classifier fuses attention evidence, aggregated prototype evidence, and maximum prototype similarities for slide-level prediction.
The framework uses only visual features and slide-level labels.

We evaluate PEAR-MIL on Camelyon16, Camelyon17, and TCGA-NSCLC using multiple feature extractors.
The main contributions are:
\begin{itemize}
  \item A dual-evidence MIL framework that retrieves prototype-indexed evidence and fuses it with an attention-derived slide feature.
  \item Evaluation across three public datasets, with parameter-matched evidence controls and prototype controls to examine the contribution of the evidence path.
\end{itemize}

\section{Related Work}
\label{sec:related}

\subsection{Pathology Foundation Models}
\label{sec:related_foundation}

Patch feature quality strongly affects WSI classification when MIL aggregators operate on frozen embeddings.
ImageNet-pretrained ResNet50 provides a general-purpose baseline \cite{he2016deep}, while pathology-specific pretraining learns representations directly from histopathology images.
CTransPath learns transferable features through contrastive pretraining \cite{wang2022transformer}, and UNI scales visual self-supervision across diverse pathology data \cite{chen2024uni}.
Related directions include image-text pretraining in CONCH \cite{lu2024visual} and tile- and slide-level pretraining in Prov-GigaPath \cite{xu2024whole}.
Even with these advances, pipelines based on frozen patch embeddings must select and combine the available features into evidence for slide-level prediction.
\method addresses this aggregation stage, motivating evaluation across feature extractors with different levels of pathology specialization, including ResNet50, CTransPath, and UNI.

\subsection{Multi-Instance Learning Approaches}
\label{sec:related_mil}

MIL methods aggregate patch features into slide-level predictions using only slide labels.
ABMIL learns gated attention weights, while CLAM adds instance-level clustering constraints \cite{ilse2018attention,lu2021CLAM}.
DSMIL combines instance- and bag-level streams, and TransMIL models patch correlations \cite{li2021dsmil,shao2021transmil}.
ACMIL and MHIM-MIL further address evidence selection through attention challenging and masked hard-instance mining \cite{zhang2024acmil,tang2023mhim}.

Prototype-based methods organize evidence around representative patterns, supporting compact slide representations and concept-level interpretation \cite{song2024morphological,sun2025prototype}.
MSPT and QPMIL-VL also use prototypes to guide patch aggregation \cite{ding2023mspt,gou2025queryable}.
TPMIL learns class- and background-associated prototypes through attention-guided assignments to refine the patch feature space \cite{yang2023tpmil}.
Building on this formulation, we investigate whether these prototypes can retrieve evidence that complements an attention-derived slide representation.
\method retrieves prototype-specific evidence tokens and fuses their aggregate with gated attention evidence for classification.

\section{Method}
\label{sec:method}

\begin{figure*}[!t]
    \centering
    \includegraphics[width=\textwidth]{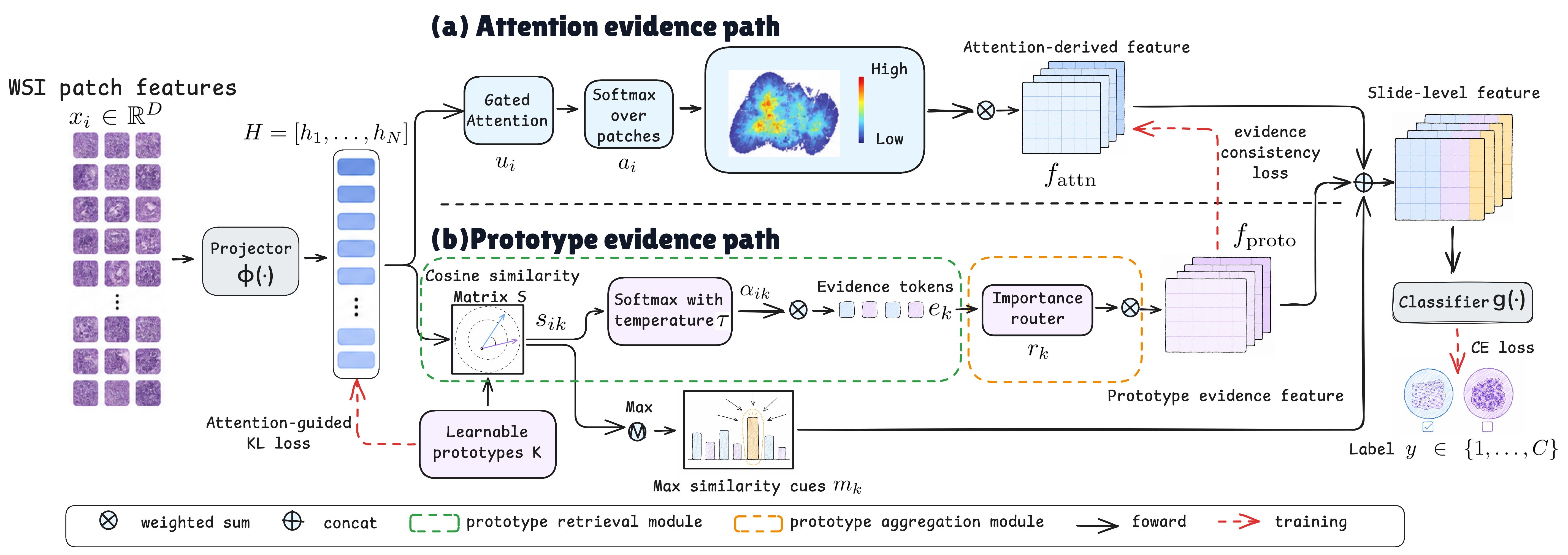}
    \caption{Overview of \method. Patch features $x_i$ are projected through $\phi$ to obtain hidden representations $h_i$. The attention evidence path (top) applies gated attention to produce $f_{\mathrm{attn}}$. In parallel, the prototype evidence path (bottom) deploys $K$ learnable prototypes $\{p_k\}$ as conditional evidence queries: each prototype computes cosine similarity with all patch features, applies temperature-scaled softmax over patches to retrieve a prototype-specific evidence token $e_k$. An evidence router aggregates $\{e_k\}$ into $f_{\mathrm{proto}}$, while the per-prototype maximum similarities $m_k$ provide additional diagnostic cues. The slide classifier fuses $[f_{\mathrm{attn}}, f_{\mathrm{proto}}, m]$ for final prediction.}
    \label{fig:method}
  \end{figure*}

An overview of the PEAR-MIL architecture is shown in Fig.~\ref{fig:method}.

\subsection{Problem Formulation}
\label{sec:method_problem}

In weakly supervised WSI classification, each slide is represented as a bag of $N$ pre-extracted patch features $X = \{x_i\}_{i=1}^{N}$ with only a slide-level label $y \in \{1,\ldots,C\}$, where $N$ varies across slides and individual patch labels are unknown.
Patch features are obtained through a standard preprocessing pipeline \cite{zhang2025accelerating,vaidya2025molecular} that performs tissue segmentation, tiles each WSI into non-overlapping patches at a fixed magnification, and encodes each patch with a frozen feature extractor.
The resulting features $x_i \in \mathbb{R}^{D}$ serve as fixed inputs to the MIL aggregator, where $D$ depends on the chosen backbone (e.g., 1024 for ResNet50, 768 for CTransPath, 1536 for UNI).

Since different feature extractors produce embeddings in different spaces, we first project all patch features into a shared hidden space through a lightweight feature projector:
\begin{equation}
  h_i = \phi(x_i), \qquad H = [h_1,\ldots,h_N]^\top \in \mathbb{R}^{N\times d},
  \label{eq:projection}
\end{equation}
where $\phi(\cdot)$ is a linear projection followed by ReLU and dropout, with hidden dimension $d=512$.
All subsequent evidence aggregation operates over the projected features $H$.

PEAR-MIL maintains $K = C+1$ trainable prototype vectors $P = \{p_k\}_{k=1}^{K}$, following TPMIL~\cite{yang2023tpmil}, comprising one prototype per class and one background prototype.
These prototypes are randomly initialized and jointly optimized with the rest of the network, serving as learnable evidence anchors that guide the retrieval of prototype-indexed diagnostic content from the patch bag.
The framework organizes slide evidence through two complementary paths that operate over the same projected features and are fused for final prediction: attention-based aggregation and prototype-conditioned retrieval.

\subsection{Attention Evidence Path}
\label{sec:method_attention}

The attention evidence path follows the gated attention mechanism~\cite{ilse2018attention}, a widely adopted module in weakly supervised WSI classification.
For each projected patch $h_i$, a patch-level importance score is computed via a gated two-branch transformation, normalized across the bag, and used to aggregate the projected features into a single slide-level vector:
\begin{IEEEeqnarray}{rCl}
  u_i &=& w^\top\bigl(\tanh(V h_i) \odot \sigma(U h_i)\bigr), \label{eq:attn_logit} \\
  a_i &=& \frac{\exp(u_i)}{\sum_{j=1}^{N}\exp(u_j)}, \label{eq:attn_norm} \\
  f_{\mathrm{attn}} &=& \sum_{i=1}^{N} a_i\,h_i, \label{eq:f_attn}
\end{IEEEeqnarray}
where $V, U \in \mathbb{R}^{L \times d}$ and $w \in \mathbb{R}^{L}$ are learnable parameters, $\odot$ denotes element-wise multiplication, and $L$ is the attention dimension.
This attention-derived feature $f_{\mathrm{attn}}$ captures diagnostically relevant global patterns and provides a strong discriminative baseline.
However, compressing the entire bag into a single aggregated vector limits the representation to one weighted summary. The prototype evidence path addresses this by introducing a structurally different, per-prototype retrieval mechanism that produces an additional, complementary slide representation.

\subsection{Prototype Evidence Retrieval and Aggregation}
\label{sec:method_evidence}

To complement the globally aggregated attention evidence, we introduce a prototype evidence path that uses learned prototype anchors to retrieve multiple evidence tokens from the same patch bag through a structurally different aggregation mechanism.
The key idea is to use each learned prototype $p_k$ as an independent evidence query: it scans all patch features, identifies content aligned with its learned direction in the feature space, and retrieves a prototype-specific evidence token.
These tokens are then aggregated into a compact representation and fused with the attention evidence for classification, as illustrated in Fig.~\ref{fig:method}.

\noindent\textbf{Evidence retrieval.}
Given the projected patch features $H$ and the prototype set $P$, we first compute the cosine similarity between every patch and every prototype, yielding a similarity matrix $S \in \mathbb{R}^{N \times K}$ with entries:
\begin{equation}
  s_{ik} = \frac{h_i^\top p_k}{\|h_i\|_2\,\|p_k\|_2}.
  \label{eq:cosine}
\end{equation}
These raw similarities are then scaled by a temperature parameter $\tau$ and normalized across patches, producing prototype-conditioned attention weights that control how strongly each patch contributes to the evidence token of prototype $k$:
\begin{IEEEeqnarray}{rCl}
  \alpha_{ik} &=& \frac{\exp(s_{ik} / \tau)}{\sum_{j=1}^{N} \exp(s_{jk} / \tau)}, \label{eq:proto_attn} \\
  e_k &=& \sum_{i=1}^{N} \alpha_{ik}\,h_i. \label{eq:e_k}
\end{IEEEeqnarray}
The temperature $\tau$ governs the retrieval sharpness: a small $\tau$ drives each prototype to attend sharply to its most similar patches, while a large $\tau$ yields a smoother, more inclusive aggregation.
Intuitively, $e_k$ represents the patch content retrieved under the similarity profile of prototype $k$, providing a prototype-indexed evidence view of the slide.

\noindent\textbf{Evidence aggregation.}
Once the $K$ evidence tokens $\{e_k\}_{k=1}^{K}$ are retrieved, they are aggregated into a compact prototype evidence representation $f_{\mathrm{proto}}$ through a lightweight evidence router.
We study two aggregation strategies: mean aggregation, which assigns uniform weights $r_k = 1/K$ to all prototypes, and importance aggregation, which learns a global per-prototype relevance logit $\rho_k$ (shared across all slides) and normalizes it via softmax.
In addition, we extract the per-prototype maximum cosine similarity as an auxiliary signal:
\begin{IEEEeqnarray}{rCl}
  r_k &=& \frac{\exp(\rho_k)}{\sum_{\ell=1}^{K} \exp(\rho_\ell)}, \label{eq:router} \\
  f_{\mathrm{proto}} &=& \sum_{k=1}^{K} r_k\,e_k, \label{eq:f_proto} \\
  m_k &=& \max_i\, s_{ik}. \label{eq:m_k}
\end{IEEEeqnarray}
The max-similarity vector $m \in \mathbb{R}^{K}$ provides the classifier with a direct, attention-free signal of the strongest presence of each prototype anywhere in the slide, complementing the aggregated evidence $f_{\mathrm{proto}}$.

\noindent\textbf{Slide-level prediction.}
The final classifier fuses the two complementary evidence views and the similarity cues into a unified prediction:
\begin{equation}
  z = g\bigl([f_{\mathrm{attn}},\, f_{\mathrm{proto}},\, m]\bigr),
  \label{eq:classifier}
\end{equation}
where $g$ is a linear layer mapping from $\mathbb{R}^{2d+K}$ to $\mathbb{R}^{C}$, and $[\cdot,\cdot,\cdot]$ denotes concatenation.
This design ensures that the classifier receives three complementary signals: (i)~the globally aggregated attention feature $f_{\mathrm{attn}}$, (ii)~the aggregated prototype evidence $f_{\mathrm{proto}}$ summarizing multiple prototype-indexed views, and (iii)~the strongest prototype-match signal $m$ per prototype.

\subsection{Training Objective}
\label{sec:method_loss}

PEAR-MIL is trained end-to-end using only slide-level labels, with a composite objective that jointly optimizes slide classification and prototype organization.

\noindent\textbf{Slide classification.}
The primary supervision comes from the standard cross-entropy loss $\mathcal{L}_{\mathrm{cls}} = \mathrm{CE}(\mathrm{softmax}(z), y)$, where $z$ is the fused prediction from (\ref{eq:classifier}).

\noindent\textbf{Prototype learning.}
To guide the prototypes toward diagnostically meaningful anchors, we adopt an attention-guided prototype objective~\cite{yang2023tpmil} that does not require any patch-level annotation.
For each patch $i$, we construct a soft prototype assignment target $q_i \in \mathbb{R}^{K}$ from the detached attention logits $u_i$ before softmax.
We first compute $b_i = (u_i - \min_j u_j)\,/\,\max(\max_j u_j - \min_j u_j,\epsilon)$ and clip it as $\tilde{u}_i = \mathrm{clip}(b_i,\delta,1-\delta)$, with $\epsilon=10^{-8}$ and $\delta=10^{-5}$.
The normalized score is distributed between the ground-truth class prototype and the background prototype: $q_{i,y}=\tilde{u}_i$, $q_{i,K}=1-\tilde{u}_i$, and $q_{i,c}=0$ for all other classes, where $K=C+1$ is the background prototype index.
Patches with high diagnostic attention are thus encouraged to be assigned to the true class prototype, while patches with low attention are pushed toward the background prototype.

Prototype prediction logits are derived from the negative normalized squared Euclidean distance between each patch and each prototype:
\begin{IEEEeqnarray}{rCl}
  \hat{q}_{ik} &=& \mathrm{softmax}_k\bigl(-\|h_i - p_k\|_2^2 \,/\, d\bigr), \label{eq:proto_logits} \\
  \mathcal{L}_{\mathrm{proto}} &=& \frac{1}{N}\sum_{i=1}^{N} \mathrm{KL}\bigl(q_i \,\|\, \hat{q}_i\bigr). \label{eq:proto_loss}
\end{IEEEeqnarray}
This loss jointly shapes the patch feature space and the prototype vectors, encouraging the prototypes to align with the attention-guided diagnostic signal.

\noindent\textbf{Evidence consistency.}
As an optional regularizer, we introduce an evidence consistency term that encourages the prototype evidence representation to align with the attention-derived feature:
\begin{equation}
  \mathcal{L}_{\mathrm{cons}} = 1 - \cos\bigl(f_{\mathrm{proto}},\, \mathrm{sg}(f_{\mathrm{attn}})\bigr),
  \label{eq:consistency}
\end{equation}
where $\mathrm{sg}(\cdot)$ denotes the stop-gradient operation, preventing the attention path from being pulled toward the prototype path.
An alternative logit-level consistency matches prototype-only and attention-anchor classifier outputs via KL divergence (studied in our ablation).

\noindent\textbf{Total objective.}
The training objective can include the following terms:
\begin{equation}
  \mathcal{L} = \lambda_{\mathrm{cls}}\,\mathcal{L}_{\mathrm{cls}} + \lambda_{\mathrm{proto}}\,\mathcal{L}_{\mathrm{proto}} + \lambda_{\mathrm{cons}}\,\mathcal{L}_{\mathrm{cons}},
  \label{eq:total_loss}
\end{equation}
where $\lambda_{\mathrm{cls}} = \lambda_{\mathrm{proto}} = 0.5$ and $\lambda_{\mathrm{cons}} = 0$ for the main results. These loss weights were selected on validation performance and kept fixed across datasets. The dual-evidence design adds only $K$ prototype vectors, $K$ router logits, and one linear fusion classifier beyond a standard attention-based MIL backbone, making it a lightweight extension that can be integrated with different feature extractors.

\section{Experiments}
\label{sec:experiments}

\subsection{Experimental Setup}
\label{sec:exp_setup}

\noindent\textbf{Datasets.}
We evaluate on three public WSI benchmarks.
\textbf{Camelyon16}~\cite{ehteshami2017diagnostic} is a lymph-node metastasis detection benchmark; we form 5 random train/validation splits from the official training set (269 slides) at an 8:2 ratio and report results on the fixed official test set (129 slides) for tumor-versus-normal classification.
\textbf{Camelyon17}~\cite{litjens20181399} provides lymph-node WSIs from multiple centers; since only the official training set (500 slides) has publicly available pN-stage labels, we restrict our experiments to these 500 slides and form 5 random patient-level train/validation/test splits at a 60:15:25 ratio to prevent data leakage across splits.
\textbf{TCGA-NSCLC}~\cite{clark2013cancer} comprises LUAD and LUSC subtyping with 1,043 slides under 5 random stratified train/validation/test splits at a 60:15:25 ratio. We use the pre-extracted UNI features released by~\cite{chen2024uni} for this dataset.

\noindent\textbf{Feature extraction.}
For Camelyon16 and Camelyon17, WSIs are processed through the Trident pipeline~\cite{zhang2025accelerating,vaidya2025molecular}, which performs tissue segmentation, tiles each slide into $256 \times 256$ patches at $20\times$ magnification, and encodes patches with a frozen feature extractor.
We experiment with three feature extractors spanning different levels of pathology specialization: ImageNet-pretrained ResNet50 (1024-d), pathology-pretrained CTransPath (768-d)~\cite{wang2022transformer}, and UNI v2 (1536-d)~\cite{chen2024uni} (referred to as UNI hereafter).

\noindent\textbf{Baselines and protocol.}
We compare PEAR-MIL against two simple pooling baselines (mean and max pooling over projected patch features) and six representative MIL aggregators: ABMIL~\cite{ilse2018attention}, CLAM~\cite{lu2021CLAM}, DSMIL~\cite{li2021dsmil}, TransMIL~\cite{shao2021transmil}, ACMIL~\cite{zhang2024acmil}, and TPMIL~\cite{yang2023tpmil}.
All methods are trained under the same optimization protocol to control for optimizer-level confounding: AdamW with learning rate $1 \times 10^{-4}$, cosine warmup over 5 epochs, weight decay $1 \times 10^{-5}$, max 100 epochs, and early stopping with patience 10 starting from epoch 40.
Every method uses the same data splits.

\noindent\textbf{PEAR-MIL configuration.}
PEAR-MIL follows the same optimization protocol as all baselines.
Unless otherwise stated, it uses hidden dimension $d = 512$, attention dimension $L = 128$, $K = C+1 = 3$ prototypes, the importance evidence router, loss weights $\lambda_{\mathrm{cls}} = \lambda_{\mathrm{proto}} = 0.5$, and $\lambda_{\mathrm{cons}} = 0$.
The evidence temperature $\tau$ is selected per backbone from a fixed candidate set $\{0.07, 0.1, 0.15, 0.3, 0.5\}$. We use the averaged validation metric over the five training splits to choose a single $\tau$ per backbone ($\tau = 0.07$ for ResNet50, $\tau = 0.3$ for CTransPath and UNI). 
All experiments report mean and standard deviation over 5 splits, with area under the ROC curve (AUC) as the primary metric.

\subsection{Main Comparison}
\label{sec:exp_main}

Table~\ref{tab:main_auc} reports the AUC and ACC comparison organized by feature extractor, including two simple pooling baselines, six representative MIL aggregators, and PEAR-MIL.
PEAR-MIL achieves the best AUC in four of the seven settings and competitive performance in the remaining three.
The most substantial gains appear in non-saturated feature regimes.
On ResNet50, PEAR-MIL reaches 84.24\% AUC on Camelyon16, improving over the strongest baseline (TransMIL, 80.39\%) by 3.85\%; on Camelyon17, TPMIL (85.04\%) and ABMIL (84.41\%) lead, while PEAR-MIL (83.75\%) shows higher variance.
With CTransPath features, PEAR-MIL achieves 95.19\% on Camelyon16 and 91.29\% on Camelyon17, both ranking first in AUC.
With UNI features, where the top baselines exceed 97.5\% AUC, performance is near saturation; PEAR-MIL remains competitive (99.23\% on C16, 98.06\% on C17) without surpassing the strongest baseline on Camelyon16 (ACMIL, 99.91\%).
The mean-pooling baseline consistently underperforms, confirming that simple feature averaging is insufficient; max-pooling is competitive on several settings but never surpasses the best attention-based or prototype-based aggregator.
The ACC results broadly follow the AUC trends, though PEAR-MIL is not consistently the top method in saturated UNI settings or in C16-ResNet50, where AUC is highest but ACC is below MaxPool and ACMIL.
Overall, the pattern is consistent with the motivation: PEAR-MIL is most valuable when patch features leave room for explicit multi-view evidence organization, and remains competitive when foundation features have already made the task highly separable. The C17-CTransPath lead over the strongest baseline is small relative to the reported standard deviations.

\begin{table*}[!t]
\renewcommand{\arraystretch}{0.96}
\centering
\footnotesize
\setlength{\tabcolsep}{3pt}
\caption{Classification results (mean $\pm$ std, \%) over five splits. The highest and second-highest mean within each backbone, dataset, and metric are \textbf{bold} and \underline{underlined}, respectively.}
\label{tab:main_auc}
\begin{tabular}{ll cccccc}
\toprule
&& \multicolumn{2}{c}{Camelyon16} & \multicolumn{2}{c}{Camelyon17} & \multicolumn{2}{c}{TCGA-NSCLC} \\
Backbone & Method & AUC (\%) & ACC (\%) & AUC (\%) & ACC (\%) & AUC (\%) & ACC (\%) \\
\midrule
ResNet50 & MeanPool & $54.71{\pm}3.18$ & $62.79{\pm}1.20$ & $63.43{\pm}8.36$ & $67.68{\pm}3.53$ & -- & -- \\
 & MaxPool & $79.80{\pm}0.76$ & $\mathbf{81.24{\pm}2.05}$ & $81.37{\pm}6.90$ & $83.36{\pm}4.03$ & -- & -- \\
 & ABMIL~\cite{ilse2018attention} & $76.64{\pm}2.65$ & $75.66{\pm}3.42$ & $\underline{84.41{\pm}4.97}$ & $\mathbf{85.28{\pm}3.38}$ & -- & -- \\
 & CLAM~\cite{lu2021CLAM} & $76.84{\pm}1.75$ & $76.12{\pm}2.05$ & $84.18{\pm}5.65$ & $\underline{84.00{\pm}4.55}$ & -- & -- \\
 & DSMIL~\cite{li2021dsmil} & $77.51{\pm}0.96$ & $74.57{\pm}0.90$ & $74.64{\pm}6.97$ & $80.00{\pm}2.99$ & -- & -- \\
 & TransMIL~\cite{shao2021transmil} & $\underline{80.39{\pm}1.93}$ & $74.88{\pm}1.87$ & $83.27{\pm}5.21$ & $82.56{\pm}3.63$ & -- & -- \\
 & ACMIL~\cite{zhang2024acmil} & $75.03{\pm}2.27$ & $\underline{80.00{\pm}2.92}$ & $82.03{\pm}7.13$ & $82.24{\pm}3.48$ & -- & -- \\
 & TPMIL~\cite{yang2023tpmil} & $76.77{\pm}1.91$ & $70.54{\pm}7.67$ & $\mathbf{85.04{\pm}5.12}$ & $\underline{84.00{\pm}4.66}$ & -- & -- \\
\cmidrule{2-8}
 & PEAR-MIL & $\mathbf{84.24{\pm}2.65}$ & $74.73{\pm}7.88$ & $83.75{\pm}7.45$ & $83.84{\pm}2.45$ & -- & -- \\
\midrule
CTransPath & MeanPool & $57.34{\pm}4.86$ & $64.19{\pm}2.42$ & $77.25{\pm}2.42$ & $73.28{\pm}2.84$ & -- & -- \\
 & MaxPool & $89.90{\pm}2.79$ & $87.13{\pm}0.79$ & $90.09{\pm}2.98$ & $86.72{\pm}4.92$ & -- & -- \\
 & ABMIL~\cite{ilse2018attention} & $91.34{\pm}4.16$ & $\underline{90.08{\pm}2.16}$ & $89.98{\pm}3.56$ & $\underline{90.40{\pm}3.08}$ & -- & -- \\
 & CLAM~\cite{lu2021CLAM} & $91.65{\pm}6.50$ & $89.61{\pm}3.59$ & $\underline{90.92{\pm}2.44}$ & $\mathbf{90.56{\pm}1.28}$ & -- & -- \\
 & DSMIL~\cite{li2021dsmil} & $93.46{\pm}3.04$ & $88.53{\pm}1.66$ & $89.78{\pm}3.05$ & $87.68{\pm}2.66$ & -- & -- \\
 & TransMIL~\cite{shao2021transmil} & $92.82{\pm}2.26$ & $89.46{\pm}2.06$ & $87.52{\pm}5.92$ & $87.04{\pm}4.39$ & -- & -- \\
 & ACMIL~\cite{zhang2024acmil} & $92.49{\pm}5.62$ & $89.30{\pm}1.73$ & $89.06{\pm}4.59$ & $88.48{\pm}3.49$ & -- & -- \\
 & TPMIL~\cite{yang2023tpmil} & $\underline{93.81{\pm}2.32}$ & $89.61{\pm}2.06$ & $89.55{\pm}4.00$ & $88.80{\pm}1.52$ & -- & -- \\
\cmidrule{2-8}
 & PEAR-MIL & $\mathbf{95.19{\pm}1.80}$ & $\mathbf{90.23{\pm}2.23}$ & $\mathbf{91.29{\pm}4.93}$ & $90.08{\pm}2.75$ & -- & -- \\
\midrule
UNI & MeanPool & $55.82{\pm}3.50$ & $59.22{\pm}5.75$ & $73.69{\pm}4.87$ & $68.80{\pm}4.77$ & $97.43{\pm}0.58$ & $91.42{\pm}1.71$ \\
 & MaxPool & $98.05{\pm}1.44$ & $91.63{\pm}2.52$ & $97.36{\pm}0.59$ & $94.08{\pm}1.09$ & $\underline{98.56{\pm}0.45}$ & $\mathbf{94.94{\pm}1.36}$ \\
 & ABMIL~\cite{ilse2018attention} & $99.64{\pm}0.23$ & $\underline{97.98{\pm}0.62}$ & $96.90{\pm}2.12$ & $94.72{\pm}1.20$ & $98.33{\pm}0.38$ & $93.95{\pm}0.82$ \\
 & CLAM~\cite{lu2021CLAM} & $\underline{99.87{\pm}0.15}$ & $\mathbf{98.29{\pm}0.31}$ & $97.67{\pm}0.82$ & $95.04{\pm}0.78$ & $\mathbf{98.59{\pm}0.41}$ & $\underline{94.79{\pm}0.79}$ \\
 & DSMIL~\cite{li2021dsmil} & $99.32{\pm}0.48$ & $93.64{\pm}2.66$ & $97.66{\pm}1.30$ & $91.36{\pm}1.55$ & $97.53{\pm}0.46$ & $93.03{\pm}1.36$ \\
 & TransMIL~\cite{shao2021transmil} & $98.74{\pm}0.62$ & $96.28{\pm}1.14$ & $96.20{\pm}1.69$ & $94.08{\pm}1.57$ & $98.32{\pm}0.64$ & $93.56{\pm}1.48$ \\
 & ACMIL~\cite{zhang2024acmil} & $\mathbf{99.91{\pm}0.08}$ & $\underline{97.98{\pm}0.93}$ & $97.14{\pm}1.22$ & $\underline{95.52{\pm}1.80}$ & $98.27{\pm}0.65$ & $93.87{\pm}1.35$ \\
 & TPMIL~\cite{yang2023tpmil} & $99.21{\pm}0.30$ & $96.59{\pm}1.87$ & $\underline{97.71{\pm}1.20}$ & $95.04{\pm}1.85$ & $98.30{\pm}0.24$ & $94.33{\pm}0.51$ \\
\cmidrule{2-8}
 & PEAR-MIL & $99.23{\pm}0.24$ & $97.36{\pm}0.62$ & $\mathbf{98.06{\pm}1.08}$ & $\mathbf{96.16{\pm}1.28}$ & $98.25{\pm}0.30$ & $93.87{\pm}1.00$ \\
\bottomrule
\end{tabular}
\end{table*}

\subsection{Ablation Studies}
\label{sec:exp_ablation}

\subsubsection{Evidence Temperature}
\label{sec:exp_tau}

\begin{figure}[tb]
  \centering
  \includegraphics[width=\linewidth]{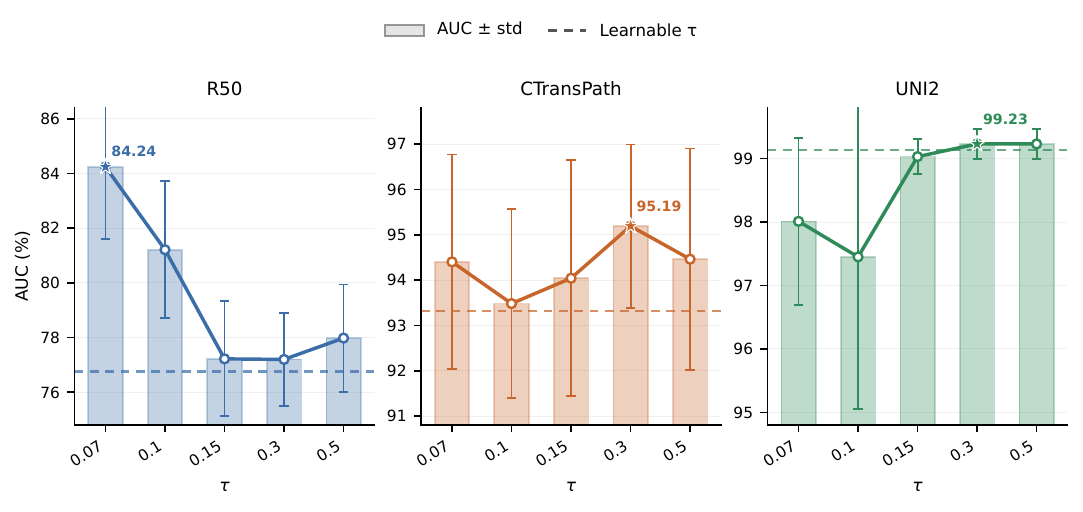}
  \caption{Temperature sensitivity on Camelyon16. AUC (\%) results across five fixed $\tau$ values ($0.07$, $0.1$, $0.15$, $0.3$, $0.5$) are evaluated over 5 splits. Dashed horizontal lines indicate the learnable-$\tau$ result for each backbone. ResNet50 peaks sharply at $\tau = 0.07$, while both CTransPath and UNI prefer smoother retrieval and achieve their best performance at $\tau = 0.3$ with flatter profiles around the optimum.}
  \label{fig:tau}
\end{figure}

The evidence temperature $\tau$ controls the retrieval sharpness of each prototype.
Fig.~\ref{fig:tau} shows the AUC of PEAR-MIL across a range of fixed temperatures and a learnable-temperature variant on Camelyon16.
The optimal temperature depends strongly on the feature extractor.
ResNet50 benefits from sharp retrieval ($\tau = 0.07$, AUC 84.24\%), with performance rapidly degrading as $\tau$ increases beyond 0.15.
CTransPath and UNI features prefer smoother retrieval ($\tau = 0.3$ and $\tau = 0.3$--$0.5$, respectively), reflecting their stronger, more semantically organized patch embeddings.
The learnable-temperature variant consistently underperforms the best fixed setting (e.g., 76.76\% vs.\ 84.24\% on ResNet50), suggesting that directly learning $\tau$ from slide-level supervision is challenging.
We therefore keep $\tau$ fixed for the main results and treat the learnable variant as a limitation.

\subsubsection{Parameter-Matched Controls}
\label{sec:exp_param}

To verify whether PEAR-MIL's gains merely reflect the enlarged classifier head, we keep the evidence classifier structure but replace prototype-indexed evidence with controlled alternatives: \textit{zero} evidence feeds a zero vector, \textit{mean} evidence feeds the average patch feature, and \textit{random} evidence feeds a deterministic random vector matched in norm to the attention feature.
Table~\ref{tab:param_controls} reports the results across three settings where full control grids are available.
On ResNet50, all three controls perform substantially below PEAR-MIL (e.g., zero evidence drops to 75.69\%, a loss of 8.55\%).
On CTransPath (both C16 and C17), the controls remain below the full model: zero and random evidence on C16-CT reach 94.14\% and 93.03\% vs.\ PEAR-MIL's 95.19\%, and on C17-CT they reach 91.19\% and 89.99\% vs.\ 91.29\%.
The consistent gap across backbones and datasets indicates that the additional classifier capacity alone cannot explain the observed improvements.

\begin{table}[tb]
\centering
\small
\setlength{\tabcolsep}{4pt}
\caption{Evidence replacement and prototype controls. AUC (\%) is reported as mean $\pm$ std. R50 and CT denote ResNet50 and CTransPath.}
\label{tab:param_controls}
\label{tab:semantic_controls}
\begin{tabular}{lccc}
\toprule
Variant & C16-R50 & C16-CT & C17-CT \\
\midrule
\multicolumn{4}{l}{\textit{Evidence replacement}} \\
Zero & $75.69{\pm}1.37$ & $94.14{\pm}1.95$ & $91.19{\pm}3.29$ \\
Mean & $77.23{\pm}1.95$ & $93.86{\pm}2.88$ & $89.85{\pm}3.86$ \\
Random & $73.74{\pm}7.88$ & $93.03{\pm}2.37$ & $89.99{\pm}4.66$ \\
\midrule
\multicolumn{4}{l}{\textit{Prototype perturbation}} \\
Random-fixed & $82.10{\pm}1.61$ & $93.47{\pm}3.15$ & $90.30{\pm}4.50$ \\
Shuffled & $83.67{\pm}1.97$ & $92.98{\pm}2.23$ & $89.91{\pm}4.20$ \\
Detached & $80.81{\pm}3.54$ & $93.38{\pm}2.44$ & $89.74{\pm}3.98$ \\
\midrule
\textbf{PEAR-MIL} & $\mathbf{84.24{\pm}2.65}$ & $\mathbf{95.19{\pm}1.80}$ & $\mathbf{91.29{\pm}4.93}$ \\
\bottomrule
\end{tabular}
\end{table}

\subsubsection{Prototype Semantic Controls}
\label{sec:exp_semantic}

We further examine whether the semantic quality of learned prototypes matters for the evidence path.
Table~\ref{tab:semantic_controls} evaluates three perturbations across three settings: \textit{random-fixed} prototypes replace learned prototypes with frozen random vectors, \textit{shuffled} prototypes rotate the learned prototype identities, and \textit{detached} prototypes preserve learned values but stop gradient from the evidence path to the prototypes.
Learned prototypes give the best AUC in all settings.
This cross-backbone comparison reveals a dual benefit that depends on feature quality.
On ResNet50, the gap between learned and shuffled prototypes is small (0.6\%), indicating that the primary gain arises from a structurally distinct aggregation mechanism (cosine-based per-prototype attention) that complements the gated attention pathway even when prototype identities are disrupted.
On CTransPath, the gap widens to 2.2\%, showing that learned prototype semantics provide additional diagnostic organization when feature quality permits.
We therefore characterize PEAR-MIL's improvement as arising from two complementary sources: a backbone-agnostic structural benefit from multi-query evidence aggregation, and a backbone-dependent semantic benefit that grows with stronger pathology-specific features.

\subsubsection{Evidence Consistency}
\label{sec:exp_consistency}

Table~\ref{tab:consistency} evaluates two optional evidence consistency variants on Camelyon16 ResNet50, both using $\lambda_{\mathrm{cons}} = 0.01$.
\textit{Feature consistency} minimizes the cosine distance between $f_{\mathrm{proto}}$ and a detached $f_{\mathrm{attn}}$; \textit{logit consistency} enforces alignment via KL divergence between prototype-only and attention-anchor classifier outputs.
Neither variant improves over the fixed-$\tau$ baseline: feature consistency achieves 81.35\% AUC (vs.\ 84.24\%) and logit consistency achieves 82.82\%.
We interpret consistency as a potential stabilizer rather than a core contributor to performance, and the main PEAR-MIL results use $\lambda_{\mathrm{cons}} = 0$.

\begin{table}[tb]
\centering
\caption{Evidence consistency on Camelyon16 ResNet50. Values are mean $\pm$ std over 5 splits.}
\label{tab:consistency}
\small
\begin{tabular}{lccc}
\toprule
Setting & AUC (\%) & ACC (\%) & F1 (\%) \\
\midrule
Fixed $\tau$ & $\mathbf{84.24{\pm}2.65}$ & $74.73{\pm}7.88$ & $73.45{\pm}6.80$ \\
Feature cons & $81.35{\pm}2.90$ & $\mathbf{75.35{\pm}3.68}$ & $\mathbf{73.66{\pm}3.15}$ \\
Logit cons & $82.82{\pm}0.97$ & $73.95{\pm}7.75$ & $72.60{\pm}6.52$ \\
\bottomrule
\end{tabular}
\end{table}

\subsection{Qualitative Evidence Inspection}
\label{sec:exp_qualitative}
\begin{figure}[tb]
  \centering
  \includegraphics[width=\linewidth]{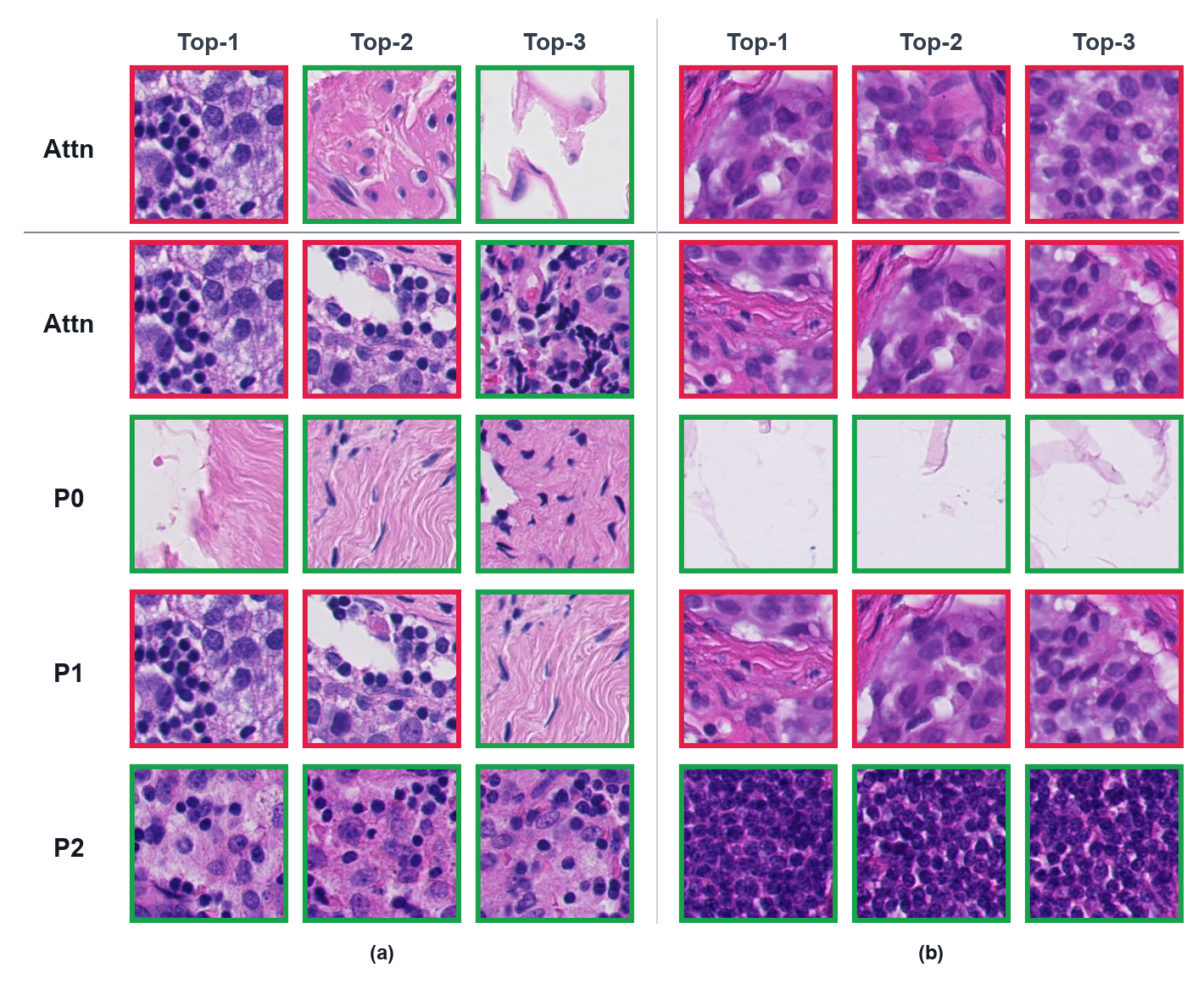}
\caption{Evidence views for Camelyon16 slides (a) \texttt{test\_110} and (b) \texttt{test\_048}, using CTransPath features, $\tau=0.3$, and split 3. Rows show ABMIL attention, PEAR-MIL attention, and prototype paths P0 (class 0), P1 (class 1), and P2 (background), respectively; columns show the top-3 patches. Red borders indicate overlap with tumor annotations, and green borders indicate no overlap. The prototype selections provide additional views of each slide.}
  \label{fig:qualitative}
\end{figure}

Fig.~\ref{fig:qualitative} shows two Camelyon16 slides (CTransPath, $\tau=0.3$) that ABMIL classifies as negative and PEAR-MIL classifies correctly as positive.
In panel (a), two of the top-3 PEAR-MIL attention patches overlap tumor annotations, compared with one for ABMIL; in panel (b), all three overlap for both methods.
The prototype paths select different patches: P1 includes tumor-overlapping patches, while the displayed P0 and P2 patches do not overlap the annotations.

\section{Discussion and Limitations}
\label{sec:limitations}

Beyond the quantitative results, several broader observations merit discussion. The prototype semantic controls reveal an interesting dependence on feature quality. This suggests that prototype-indexed evidence is best understood as a dual-benefit mechanism whose dominant gain factor shifts with backbone capability, an insight that may inform how prototype-based methods are designed and evaluated. The temperature sensitivity pattern, where weaker features benefit from sharp retrieval and stronger ones from smooth retrieval, highlights a practical tension between retrieval specificity and feature quality that warrants further study.

Several limitations also point to concrete directions for future work. The learnable-$\tau$ variant underperforms fixed temperatures, and we attribute this to the fact that the cross-entropy loss provides only an indirect signal for temperature learning; identifying a more suitable auxiliary objective or validation-driven adaptation schedule remains an open challenge. The current evidence router employs global prototype weights shared across all slides; conditioning these weights on per-slide content is a natural next step that may improve adaptation to heterogeneous tissue patterns. On the evaluation side, extending the dual-evidence framework beyond binary classification to tasks such as multi-class subtyping and survival prediction would further test its generality. More broadly, systematic prototype-diversity and localization studies across larger cohorts would strengthen the interpretability claims suggested by our qualitative examples and semantic controls.

\section{Conclusion}
\label{sec:conclusion}

We presented PEAR-MIL, a Prototype Evidence Aggregation and Retrieval framework for weakly supervised WSI classification in which learned prototypes serve as conditional evidence queries.
Each prototype retrieves a prototype-specific evidence token from the patch bag through similarity-based matching; these tokens are aggregated into a compact prototype evidence representation and fused with an attention-derived slide feature to provide complementary evidence views for slide-level prediction.
The framework is lightweight, purely visual, and trained end-to-end using only slide-level labels.

Experiments on Camelyon16, Camelyon17, and TCGA-NSCLC across ResNet50, CTransPath, and UNI backbones show that PEAR-MIL achieves state-of-the-art or competitive performance, with the clearest gains in non-saturated feature regimes.
Parameter-matched controls indicate that these gains are not explained by additional classifier capacity. Prototype semantic controls further show that the improvement arises from a structurally distinct evidence aggregation mechanism (cosine-based per-prototype attention) that complements gated attention, with learned prototype semantics providing additional benefit when feature quality permits.

\bibliographystyle{IEEEtran}
\bibliography{bibtex/bib/bibm_official}
\end{document}